# Chapter 19:
# Bots against Bias: Critical Next Steps for Human–Robot Interaction

**KATIE SEABORN**

*Industrial Engineering & Economics*
*Institute of Science Tokyo*

**ABSTRACT:** *We humans are biased – and our robotic creations are biased, too. Bias is a natural phenomenon that drives our perceptions and behavior, including when it comes to socially expressive robots that have humanlike features. Recognizing that we embed bias, knowingly or not, within the design of such robots is crucial to studying its implications for people in modern societies. In this chapter, I consider the multifaceted question of bias in the context of humanoid, AI-enabled, and expressive social robots: Where does bias arise, what does it look like, and what can (or should) we do about it. I offer observations on human–robot interaction (HRI) along two parallel tracks: (1) robots designed in bias-conscious ways and (2) robots that may help us tackle bias in the human world. I outline a curated selection of cases for each track drawn from the latest HRI research and positioned against social, legal, and ethical factors. I also propose a set of critical next steps to tackle the challenges and opportunities on bias within HRI research and practice.*

## 19.1    Introduction

Bias, whether toward humans or humanoid robots, is a natural function of the human brain.1 We have evolved brains that are able to craft models of the world and then use these models to make rapid judgments when faced with new information. This can be, and perhaps in the more distant past often was, a matter of survival under harsh, dynamic, and unpredictable conditions. But it is also a fundamental cognitive tool for daily life now. Decision-making happens at all scales, from the insignificant to the highest of stakes. Decisions can involve a complex array of shifting factors or a simple few.2 Time pressure and time scales can vary. Some require concerted and conscious effort, while others are best served by the speed of our unconscious mind. Sometimes we do not have a choice in the matter. Indeed, as Kahneman, Tversky, and others have persuasively demonstrated across decades of work, our brains opt to think quickly and less rigorously, unless directed otherwise. From an evolutionary perspective, this may be grounded in our survival as a species.3 But what is then directing our brains when it is not our conscious minds? One answer is bias. While much of the above discussion applies to humans, I propose that it equally applies to robots, especially those that are humanoid in appearance, AI-enabled, expressive, and interacting with humans in social contexts. For such robots, bias is an especially timely topic and raises provocative questions for law, policy, and ethics.

Bias refers to a cognitive inclination toward one of several options. Bias on its own is not necessarily unconscious or valenced. Nevertheless, we tend to focus on when bias does harm without us realizing it. Welsh and Begg,3 for instance, define bias as the "unconscious errors resulting from how people's minds work." Furthermore, bias emerges during decision-making, when we deploy a heuristic or practical method to solve a problem.4 It can be unconscious (implicit) or conscious (explicit) in orientation. It can take place in the brain (cognitive bias), be observed in behavior (prejudice), and be embedded in the things that we create, including machines (statistical and data). Tversky, Kahneman, Ariely, Thaler, and other seminal researchers[5] have spent decades identifying biases, determining when and where they appear, and determining the extent of their influence. Common examples include overconfidence bias ("I can definitely do this, even though I've only done it once … when I was a kid!"), confirmation bias ("this must be true because I already believe it"), anchoring ("the color of his eyes is incredible … I can vote for this candidate with confidence"), risk aversion ("I'm not happy with my lot but that sounds too difficult to try instead"), and stereotyping ("women love chocolate"), to name a few.

---

1 Eberhardt, J. L., Biased: Uncovering the Hidden Prejudice That Shapes What We See, Think, and Do, Penguin, 2020.

2 Tversky, A. and Kahneman, D., Judgment under Uncertainty: Heuristics and Biases: Biases in Judgments Reveal Some Heuristics of Thinking under Uncertainty. Science 185(4157), 1124–1131, 1974. www.science.org/doi/abs/10.1126/science.185.4157.1124; Haselton, M. G., Nettle, D., and Andrews, P. W., The Evolution of Cognitive Bias. The Handbook of Evolutionary Psychology 724–746, John Wiley & Sons: New York, NY, USA. 2015.

3 Welsh, M. and Begg, S., What Have We Learned? Insights from a Decade of Bias Research. The APPEA Journal 56(1), 435–450, 2016. www.publish.csiro.au/aj/AJ15032.

4 Tversky & Kahneman (1974), supra note 2, at 1; Kahneman, D., Slovic, S. P., Slovic, P., and Tversky, A. (eds.), Judgment under Uncertainty: Heuristics and Biases, Cambridge University Press: Cambridge, UK, 1982.

Understanding how bias occurs with AI-powered humanoid and socially expressive robots requires a closer look at a few intertwined concepts. Bias of the cognitive variety is a psychological construct, made up of cognitive associations that model the world and influence what we do in it. Many of these are social constructs, based on shared and collaboratively developed ideas accepted by a group of people, including whole cultures and societies. Gender, for instance, refers to the roles, modes of expression, and norms that a society has deemed appropriate for people with a certain cluster of characteristics generally traced back to biological sex. Bias can lead us to apply these models to robots, where we perceive gender in a robot and then act in gender stereotypical ways toward the robot. Similarly, race and ethnicity are human classification systems premised on certain physical traits; for example, skin color, shared cultural qualities, and ancestry. Such constructs make up the mental models we have of the world and guide reactions and decision-making. Mental models are a representation of reality, built up over time from our personal experience as embodied beings in the world. When we encounter robots shaped in our image, even with the slightest humanlike cues of gender, race, cultural background, age, and so on, we draw on these mental models, reflexively and often without realizing it. This is because biases rely on these models to occur. Biases influence our attitudes, the evaluative cognitive judgments we have about certain objects, and our behavior, the actions we take in the world. Indeed, reflecting on our thinking and examining our actions can reveal what biases are or have been at play. We are not always primed to do so, however, which is where human-interfacing robots may be able to step in.

---

[5] For example: Tversky, A. and Kahneman, D., Availability: A Heuristic for Judging Frequency and Probability. Cognitive Psychology 5(2), 207–232, 1973. https://doi.org/10.1016/0010-0285(73)90033-9; Kahneman, D. and Tversky, A., Intuitive Prediction: Biases and Corrective Procedures, Decisions and Designs, Incorporated. No. ADA047747, 1977. https://apps.dtic.mil/sti/citations/ADA047747; Kahneman, D. and Tversky, A., The Simulation Heuristic, 1981. No. ADA099504; De Bondt, W. F. and Thaler, R., Does the Stock Market Overreact? The Journal of Finance 40(3), 793–805, 1985. https://doi.org/10.1111/j.1540-6261.1985.tb05004.x; Kahneman, D., Knetsch, J. L., and Thaler, R. H., Anomalies: The Endowment Effect, Loss Aversion, and Status Quo Bias. Journal of Economic Perspectives 5(1), 193–206, 1991. https://doi.org/10.1257/jep.5.1.193; Tversky, A. and Kahneman, D., Loss Aversion in Riskless Choice: A Reference-Dependent Model. The Quarterly Journal of Economics 106(4), 1039–1061, 1991. https://doi.org/10.2307/2937956; Ariely, D., Loewenstein, G., and Prelec, D., "Coherent Arbitrariness": Stable Demand Curves Without Stable Preferences. The Quarterly Journal of Economics 118(1), 73–106, 2003. https://doi.org/10.1162/00335530360535153.

Few would argue against the existence of bias and its significance for social robotics research and practice. At the same time, many of us still struggle to address bias conscientiously and productively in our work. For instance, new evidence suggests that how we write up our work provides subtle but unmistakable cues about our own biases, at least when it comes to gender.6 We have the gist and some pertinent examples, but not the depth of understanding, training and education, or resources needed to effect change or explore bias in greater depth. Yet, we are moving in this direction, as recent panels and workshops,7 calls for papers,8 grants,9 and other forms of academic engagement10 have signposted. Initiatives aiming for team diversity as a solution have appeared. For example, Black in Robotics (BiR)11 was launched to address entrenched inequalities in who roboticists are, what issues we feel are important, and providing the impetus for making a change in robotics and research practice. We are interested and change is on the horizon – we just need a path to get there. Two avenues are available to us. First, we need to more deeply understand bias as it plays out within robotics practice: When does it happen, and under what circumstances? What does it look like in robots, in research, within us? And what can (or should) we do to address it? Second, and more provocatively, we can harness our knowledge of biases in productive ways. Can we embed biases in the design of interactions between people and robots as a form of training or just-in-time support? Can robots make use of the biases we tend to fall prey to so as to encourage us when we are engaging in difficult but beneficial activities? Can robots protect us, shift our behavior, or inspire us?

---

6 Seaborn, K. and Frank, A., What pronouns for Pepper? A critical review of gender/ing in research. Proceedings of the 2022 CHI Conference on Human Factors in Computing Systems, 1–15, 2022. https://doi.org/10.1145/3491102.3501996.

7 Including, but not limited to: Workshop on Roboethics at the 2009 IEEE International Conference on Robotics and Automation (ICRA 2009); A quest for Avoiding Harmful Bias and Discrimination by robots (AGAINST-19) at ICRA 2019; Bots Against Bias (BoAB 2021) at the 2021 IEEE International Symposium on Robot and Human Interactive Communication (RO-MAN 2021); GENDERING ROBOTS: Intersectional Feminist Perspectives on Gender in Social Robotics (GenR Workshop 2022) at RO-MAN 2022; and the Diversity, Equity, and Inclusion (DEI) Inclusive HRI Workshop at the ACM/IEEE International Conference on Human-Robot Interaction (HRI '22).

8 For example, the 2021 call for critical robots research (AI & Society) and the 2022 call for GENDERING ROBOTS: Ongoing (Re)configurations of Gender in Robotics (International Journal of Social Robotics).

9 For example, the 2021 call for "Tackling gender, race and other biases in AI" in the HORIZON programme by the European Commission.

10 Work at the intersection of industry and academia may be relevant. Take, for instance, natural language processing (NLP) debiasing work conducted by researchers and practitioners at Amazon for Alexa and published at the 2020 iteration of the Empirical Methods in Natural Language Processing (EMNLP 2020) conference, for which Amazon is a "diamond" sponsor.

11 https://blackinrobotics.org

Answering these questions will require a deeper, more critical form of engagement that is driven by multidisciplinary fertilization and reflexivity, as well as sensitivity to the specifics of social robotics and HRI research and practice. To this end, I offer a critical narrative review of HRI research covering two entwined tracks: bias as a challenge to be addressed in robots and HRI research, and bias as a challenge that robots and HRI research can address. I aim to raise awareness, provoke corrective changes in social robotics and HRI research, and spark innovation within robotics praxis. I aspire to recognize how unexamined biases have affected our work while at the same time demonstrating the potential of robotics to address these very biases. On the one hand, robots are tools, machines with sensors, actuators, and access to external data sources. On the other hand, robots are interactive agents, expressive and modelled after we humans to the extent that they can move autonomously, make decisions, and communicate with us. As Perugia et al.[12] suggest, robots are created in "the shape of our biases," but as they are our creations, it is within our power to intentionally shape them and what influence they have on our world.

This chapter is structured on two tracks: robots against bias and against bias in robots. For each, I provide a curated selection of case studies that have academic, industrial, and societal merit. I cover examples from seminal research and the latest work. I outline critical next steps alongside the social, legal, and ethical factors that may be relevant to informing policy and regulations within industry and academia. I do not attempt to be all encompassing or prescriptive. Rather, I aim to incite reflection, prompt ideation, and set the stage for action, gearing the content for scholars, engineers, and practitioners working with modern humanoid robots deployed as social agents among people. This is not a solo effort; it is a community effort. My goal is to strike a chord across the social robotics, HRI, and adjacent communities – one that is clear, actionable, and inspiring.

---

[12] Perugia, G., Guidi, S., Bicchi, M., and Parlangeli, O., The Shape of our Bias: Perceived Age and Gender in the Humanoid Robots of the ABOT Database. In Proceedings of the 2022 ACM/IEEE International Conference on Human-Robot Interaction, pp. 110–119, March 2022. https://dl.acm.org/doi/10.5555/3523760.3523779.

## 19.2 Track: Robots against Bias

We are at the mercy of our evolutionary history.13 Many of the cognitive shortcuts and biases we host developed in our ancestors as successful responses to a wild and hostile world. Now the world is smaller, a direct consequence of human mastery over nature. Globalization, world travel, telecommunications, and especially the internet have all changed the way we relate to our world, as well as created new worlds never before experienced by any species on earth. These technological worlds are nevertheless processed by human brains. Evolution happens at a slow pace and may largely be outside of our hands, at least directly. Nevertheless, we have full control over our technological world crafting. Can we create robots that help us become aware of, if not overcome, cognitive pitfalls and logical fails when appropriate? Can robots scaffold our thinking? Can they save us from ourselves, and does everyone want to be saved – in the same way? I offer several cases where robots could intervene when human processing fails in prosocial, ethical ways.

### 19.2.1 CASE: MANIPULATION

Reality may not always be what it seems. Sometimes this is a matter of subjectivity. Other times it is a matter of circumstance. And still other times it can be a matter of deception. People or their agents may purposefully distort information about or representations of reality. I provide two pertinent examples of where AI-powered and socially aware robots can intervene: to assess our resilience against misinformation in the future infodemic 2.0 and as faithful interviewers during sensitive situations where being human may be a deficit.

#### 19.2.1.1 *Example: Robots That Simulate Our Dystopian Futures*

We are surrounded by opportunities to be subjected to fake news. Every time we log onto social media or even check in with our friends, family, and acquaintances online, we are at risk of exposure. AI-enabled humanoid robots, as well, can be a source of deception and false news when we engage them for information-seeking tasks. Misinformation is not necessarily obvious. A survey in 2022 found that between 45 percent and 55 percent of respondents from around the world noticed fake news ... and this was only what they noticed.14 Bots that shill fake news are sophisticated, mimicking conventional speech and echoing the mantras of the ideologies that they represent. We are no longer good at detecting real people from fake accounts. We are also at risk when people we trust fall prey to misinformation. Misinformation spreads through social networks based on human relationships. Like dominos, it only takes one to fall for the rest to cascade down the line. Falling prey to fake news is embarrassing, but it is also a matter of social responsibility, ethics, and law. Believing in and spreading fake news has had world-shaking implications for global matters, from the 2017 US election[15] to vaccine controversies and mask mandates during the COVID-19 pandemic.[16] Attempts to raise awareness and curtail the spread of misinformation are in progress. Emerging research offers a mixed picture on the effectiveness of known protective, preventative, and post-hoc measures. Such a multifaceted, dynamic, and widespread problem is likely to need a multipronged approach. One prong may be robotic interventions.

---

13 Kahneman, D., Thinking, Fast and Slow, Macmillan: London, UK, 2011.

14 www.statista.com/statistics/1317019/false-information-topics-worldwide

What havoc could self-determined, self-replicating, and uber-connected social media bots have on the spread of fake news? Arguably, we are already living in such a world. With this in mind, Wang, Angarita, and Renna[17] extrapolate the fake news purveying bots of the present day into an extreme future. They offer a model of "social bots" that have evolved within social networks to the extent of being able to create, gather, and spread fake news ... and have a taste for it. This vision relies on a model of information diffusion conceptualized by Myers, Zhu, and Leskovec.[18] This notion of information diffusion, inspired by biological viral models and metaphors, reflects actual information spread from a month's worth of Twitter data. Exposure is not simply a matter of spread through social networks, but also external factors beyond the network – roughly one third of the infection rate. What these external factors are is ambiguous. They could be other websites, nonnetworked social interactions (e.g., conversations with friends and family IRL), individual characteristics, ... confounding factors in the wild are hard to identify. Nevertheless, Wang, Angarita, and Renna foretell that present-day attitudes and behaviors in social media contexts fueled by human cognitive biases will increase the effectiveness of these viral misinformation agents. They imagine swarms and colonies of social bots working in sync to mete out misinformation attacks on unsuspecting social media users. All of this is technologically feasible and may even be live right now in a nascent form. What remains to discover is the size of this near-future problem. Are humans more resilient than models would predict? We can test this out now with computer simulations and real users. We must also consider how virtual characters and social robots could enter the picture, as human-accessible embodiments of these swarms. Virtual influencers, for instance, are a modern phenomenon on social media and social networking services (SNS).19 These influencers are not people, even if they appear humanlike in appearance and behavior, yet they are followed by thousands, if not millions, of people (or at least accounts). At present, most virtual influencers appear to be human powered or human supervised rather than AI powered, but autonomous and socially expressive versions are on the horizon. It is just a matter of time.

---

15 Bovet, A. and Makse, H. A., Influence of Fake News in Twitter during the 2016 US Presidential Election. Nature Communications 10(1), 1–14, 2019. https://doi.org/10.1038/s41467-018-07761-2.

16 Loomba, S., de Figueiredo, A., Piatek, S. J., de Graaf, K., and Larson, H. J., Measuring the Impact of COVID-19 Vaccine Misinformation on Vaccination Intent in the UK and USA. Nature Human Behavior 5(3), 337–348, 2021. https://doi.org/10.1038/s41562-021-01056-1.

17 Wang, P., Angarita, R., and Renna, I., Is this the Era of Misinformation Yet: Combining Social Bots and Fake News to Deceive the Masses. In Companion Proceedings of the Web Conference 2018, pp. 1557–1561, April 2018. https://doi.org/10.1145/3184558.3191610.

18 Myers, S. A., Zhu, C., and Leskovec, J., Information Diffusion and External Influence in Networks. In Proceedings of the 18th ACM SIGKDD International Conference on Knowledge Discovery and Data Mining, pp. 33–41, August 2012. https://doi.org/10.1145/2339530.2339540.

From the above, I offer the following as factors to consider when exploring how bias in the embodiments of robots and other AI simulate, or could simulate, our future information highways.

**Critical next steps:**

- Computer simulations of extreme misinformation bots, building on the work of Wang, Angarita, and Renna.

**Social factors:**

- Field explorations of the prevalence of swarms in social media and social networks, and what social influence these swarms are having.

**Legal factors:**

- Governance of social media platforms is needed. What this may look like could vary, but is likely a community effort involving governments, regulating bodies, the social network companies, nonprofit and neutral third parties, and the people using the platforms.
- Recognition and treatment of social bots as legal agents rather than neutral technologies should be considered.

**Ethical factors:**

- Philosophical and governance engagement on the question of who determines what is fake and what is real news and how this decision is made.
- Moreover, engagement on the question of who is responsible for harms resulting from deception and misinformation attacks: the authors, even if they are bots? The platforms that host this material and provide the means by which it spreads? The governments who offer no intervening laws? Or no one, because it is our own responsibility to decide for ourselves?
- Finally, questioning whether human creators are responsible after social bots become autonomous and self-replicating, and if not, who or what is responsible for the behavior of these autonomous bots.

---

19 Hsu, T., These Influencers Aren't Flesh and Blood, Yet Millions Follow Them. New York Times, 2019. www.nytimes.com/2019/06/17/business/media/miquela-virtual-influencer.html.

### 19.2.1.2  *Example: Harnessing Robots That Fail to Mislead People*

Deceit is a human phenomenon. People lie, hold back the truth, manipulate facts, mislead, and misdirect. Reasons run the gamut from self-interest to protecting others to malicious delight … to simple bias. When a crime happens and we must rely on witness testimony, bias can come into play. A range of cognitive pitfalls and biases have been identified on the side of the witness as well as the interviewer.[20] Memory is fallible: even if we intend to accurately portray the truth as we know it, our brains are incapable of faithfully recording every detail of an event.[21] In fact, recall is noisy, with information loss every time a memory is brought to mind.[22] We are also not good at remembering details over the gist of an event or scene.[23] A car may have been red or maroon or even blue. Witnesses may also be keen to help or feel pressure under the guidance of an expert who is interviewing them, leading to a range of well-mapped biases. Social acceptability bias, for instance, may lead the witness to withhold, massage, or outright lie about what they remember.[24] At least, when faced with a human interviewer.

---

[20] For instance: Brigham, J. C., Maass, A., Snyder, L. D., and Spaulding, K., Accuracy of Eyewitness Identification in a Field Setting. Journal of Personality and Social Psychology 42(4), 673, 1982. https://psycnet.apa.org/doi/10.1037/0022-3514.42.4.673; Murrie, D. C. and Boccaccini, M. T., Adversarial Allegiance among Expert Witnesses. Annual Review of Law and Social Science 11, 37–55, 2015. https://doi.org/10.1146/annurev-lawsocsci-120814-121714.

[21] Howe, M. L. and Knott, L. M. The Fallibility of Memory in Judicial Processes: Lessons from the Past and their Modern Consequences. Memory 23(5), 633–656, 2015. https://doi.org/10.1080/09658211.2015.1010709.

[22] Baddeley, A. D., Human Memory: Theory and Practice, Psychology Press, 1997.

[23] Friedman, A., Framing Pictures: The Role of Knowledge in Automatized Encoding and Memory for Gist. Journal of Experimental Psychology: General 108(3), 316, 1979. https://psycnet.apa.org/doi/10.1037/0096-3445.108.3.316.

[24] For a literature review, refer to Krumpal, I., Determinants of Social Desirability Bias in Sensitive Surveys: A Literature Review. Quality & Quantity 47(4), 2025–2047, 2013. https://doi.org/10.1007/s11135-011-9640-9.

Social and humanoid robots, by virtue of not being human but humanlike enough to communicate, may offer a neutral alternative. Bethel et al.25 explored whether a robot or human was more or less influential when in the role of an interviewer seeking out eyewitness testimony on a fabricated crime. Some participants were given misleading information while others were given information that matched the details of the purported crime. After a series of distractor tasks, and time, participants were asked to report on what they remembered. Significantly, the human interviewer was found to mislead participants to a greater degree than the duplicitous robot interviewer. However, the authors warn that these results may be tied to the less-than-favorable perceptions that people had about the robot, thus disrupting its social influence.

Human interviewers in sensitive roles are likely to be trained and experienced. Nevertheless, bias can creep in. Interviewers may expect a certain result or have a stake in a certain outcome, or even be influenced by unrelated factors. For example, at the end of a long day the interviewer may feel tired, stressed, or rushed. More subtly, the interviewer may have internalized biases against social categories such as gender, race, accent, economic status, and social status (i.e., celebrities).26 And, like interviewees, interviewers may misremember the details of a case. On the other hand, interviewees may fall prey to the authority bias when faced with information from a trusted, expert source that conflicts with what they remember.27 In the face of authority, participants may comply with the flow set by the expert.28 Robots, despite being humanlike in form and communication, may be perceived as socially neutral or morally unaccountable.29 They may thus be ideal in cases where an unimpeded account of a sensitive event is needed. When the honest truth about a tough situation is needed, a robot may be the best choice. I offer the following, inspired by the above

---

25 Bethel, C. L., Eakin, D., Anreddy, S., Stuart, J. K., and Carruth, D., Eyewitnesses are misled by human but not robot interviewers. In 2013 8th ACM/IEEE International Conference on Human-Robot Interaction (HRI), pp. 25–32, March 2013, IEEE. https://dl.acm.org/doi/pdf/10.5555/2447556.2447562.

26 Refer to the literature on job interview and resume biases, such as Purkiss, S. L. S., Perrewé, P. L., Gillespie, T. L., Mayes, B. T., and Ferris, G. R., Implicit Sources of Bias in Employment Interview Judgments and Decisions. Organizational Behavior and Human Decision Processes 101(2), 152–167, 2006. https://doi.org/10.1016/j.obhdp.2006.06.005.

27 Milgram, S., Behavioral Study of Obedience. The Journal of Abnormal and Social Psychology 67(4), 371, 1963. https://psycnet.apa.org/doi/10.1037/h0040525.

28 Cialdini, R. B. and Goldstein, N. J., Social Influence: Compliance and Conformity. Annual Review of Psychology 55(1), 591–621, 2004. https://doi.org/10.1146/annurev.psych.55.090902.142015.

29 Kahn Jr, P. H., Kanda, T., Ishiguro, H., Freier, N. G., Severson, R. L., Gill, B. T., ... and Shen, S., "Robovie, You'll Have to go into the Closet Now": Children's Social and Moral Relationships with a Humanoid Robot. Developmental Psychology 48(2), 303, 2012. https://psycnet.apa.org/doi/10.1037/a0027033.

literature, as explorations on how robots may or may not be harnessed to lead, or even mislead, people:

**Critical next steps:**

- In-court robot interviewers can be explored in the wild, in real court situations.
- Remote human-in-the-loop interviewers with robot bodies; that is, telerobot interviewers, may experience a "social influence disruption" on account of having robotic embodiments, but this needs to be explored.

**Social factors:**

- Social biases and social identity theory may help explain and predict when and how social influence can be elicited or disrupted.

**Legal factors:**

- We may consider legislating robots as bodies (controlled by remote humans) and/or agents deployed to gather eyewitness testimony.
- We may need to consider governance of interview practices and questions as well as how to manage sensitive answers and ensure data integrity.
- Follow-up governance of data analysis and use will be needed.

**Ethical factors:**

- Duplicity in the case of human-in-the-loop robotic avatars is an ethical concern.
- Possible overtrust and unexpected disclosures may occur. We must also consider how such reactions may impact the interviewee and the people governing the robot and/or the data.

### 19.2.2   CASE: METACOGNITION

We cannot always avoid being manipulated, influenced, persuaded, or otherwise falling prey to bias. So, what can we do? Perhaps living with bias is a matter of being aware of it. And perhaps robots can step in when we encounter bias in ourselves or within a manipulative scenario, boosting our performance in terms of recognizing, understanding, and consciously reacting to murky situations. Metacognition is an umbrella concept referring to a top-down procedure of monitoring and controlling our cognition, commonly summarized as "thinking about our thinking."30 Several models describing the process of metacognition exist. Nelson and Narens31 framed metacognition as a matter of monitoring and control over memory activities, specifically acquisition, retention, and retrieval related to information and subsequently our modeling of that information. Efklides32 proposed an underlying three-part structure of metacognition made up of metacognitive knowledge, or the knowledge and information we have about thinking, metacognitive experience, or the active awareness of how we think, feel, and are oriented toward thinking during cognitive tasks, and metacognitive skills, or the strategies that we purposefully deploy when thinking consciously. Expanding on the Nelson–Narens model with a learning-centric frame, Dunlosky and Metcalfe33 operationalized metacognition as a three-part, cyclical activity made up of planning for meta-level thinking, monitoring one's performance, and reflecting on one's ability to be aware of how one performed metacognitively. All of these are human activities that can help us think better, or at least be more aware. Can we automate these activities or strategies or extend our existing capacities with AI-based robots or robotic implements? I highlight three ways in which robots may be designed to offer "teachable moments," intervene when we need a metacognitive boost, and even capitalize on our own tendency to fall prey to various biases, for our own good.

---

30 Flavell, J. H., Metacognition and Cognitive Monitoring: A New Area of Cognitive–Developmental Inquiry. American Psychologist 34(10), 906, 1979. https://psycnet.apa.org/doi/10.1037/0003-066X.34.10.906; Jameson, K.

### 19.2.2.1  *Example: "Teachable Moments" with Robots That Fall Prey to Biases*

One way to learn is to help others learn. A robot could be your pupil in metacognition. When the robot makes a logical misstep, you can intervene – and thereby increase your own performance when it comes to similar mishaps. But will you know when to intervene and how? Along these lines, Biswas and Murray34 designed three robots, ERWIN, MyKeepon, and MARC, to fall prey to five kinds of cognitive biases: misattribution bias, where people incorrectly attribute a cause to an event or link between events that does not exist; empathy gaps, when one is feeling a strong emotion and is unable to imagine being in another emotional state, such

---

as when one is angry and cannot imagine being calm; the Dunning–Kruger effect,35 where people overestimate their ability; self-serving bias, when people assign themselves as the cause of good things that happen to them and external factors when bad things happen to them; and the humor effect, where people are more likely to remember an event if it is humorous. The robots were deployed in long-term settings as conversational and entertainment companions, ideal settings for these biases that allowed for identification of longitudinal results. Although the robots differed across studies, the result was the same: biased robots were better. Specifically, people liked the robots that were imperfect more than those that were not.

The authors suggest that such biases can pave the way for long-term engagement with robots in daily life. However, the ability of participants to identify and respond to biases embedded in the design of interactions with these robots suggests another opportunity: learning about these very biases. As Azevdeo36 suggests, artificial agents, from social robots to voice assistants to virtual characters and beyond, may be an effective way to safely and consistently engage people in metacognitive activities such as detection of cognitive biases. We can also draw from the serious games literature. For instance, Basol, Roozenbeek, and van der Linden37 created the Bad News game, which teaches people how to make fake news … on purpose. This gamified intervention draws inspiration from the notion of vaccines as an inoculation strategy against misinformation, whereby people are exposed to small, repeated doses of the mechanisms by which fake news works. Over time, people get better and more confident in their assessments of real fake news. A companion robot that is clearly under the influence of misinformation could be an effective complement to such interventions. By helping your robot companion understand what cognitive biases and heuristics underlie its reasoning, you may in fact be helping yourself prepare for the real world.

I now offer potential "teaching moments" and related provocations for further exploration:

**Critical next steps:**

- We can explore robots that fall prey to biases and need our help to become aware of them as a means of teaching about bias.

---

35 Dunning, D., The Dunning–Kruger Effect: On being Ignorant of One's Own Ignorance. In Advances in Experimental Social Psychology, Academic Press, Vol. 44, pp. 247–296, 2011. https://doi.org/10.1016/B978-0-12-385522-0.00005-6.

36 Azevedo, R., Reflections on the Field of Metacognition: Issues, Challenges, and Opportunities. Metacognition and Learning 15(2), 91–98, 2020. https://doi.org/10.1007/s11409-020-09231-x.

37 Basol, M., Roozenbeek, J., and van der Linden, S., Good News about Bad News: Gamified Inoculation Boosts Confidence and Cognitive Immunity against Fake News. Journal of Cognition 3(1), 2020. https://doi.org/10.5334%2Fjoc.91.

- We can explore robots that are intentionally contrary and/or disruptive in contextually sensitive ways, such as within social media spaces.
- We can explore robots that incite cognitive dissonance and thereby elicit an opportunity to practice metacognition by way of thinking about this dissonance.

**Social factors:**

- We need to determine that the effects are the "right" way – that robots are teaching us to be aware of bias and not teaching us how to be biased or manipulate others.

**Legal factors:**

- Customized robots that link negative biases to individuals need to be carefully secured, as this data may constitute sensitive information about personal weaknesses that others could exploit for monetary or other gains.

**Ethical factors:**

- Uncontrolled settings, such as longitudinal in-the-wild deployments, will need careful design and monitoring to ensure that teachable moments do not backfire.
- Users need to have access to and control over the sensitive data guiding the teachable moments, including the ability to opt out and erase this data.

### 19.2.2.2  Example: Robots That Intervene When We Fail

Every time we engage with the digital world, we must make decisions about the information we receive and should accept that we are at risk of deception. What if robots were there by our side, boosting our abilities or intervening before we fall prey to known tactics? Robots could be designed as ever-present aids or cognition extenders, ready to help us on a moment's notice. How can robots intervene? Kozyreva, Lewandowsky, and Hertwig[38] recently proposed a general three-part framework of strategies that could be applied to the design of robots as aids when we participate in uncertain and potentially hostile digital environments. These strategies support our control over our own cognition and behavior and are founded on decades of experimental research. Nudging[39] involves manipulating the choice architecture or what choices are available to us, especially those that are closest and easiest and most desirable to us, in our favor. Boosting[40] are strategies that aim to improve our cognitive and motivational competencies to enhance our abilities and behavior. Technocognition[41] takes our knowledge of people's psychology and applies it to the design of technological solutions. In a sense, any robotic intervention could be a form of technocognition. Then, we may consider two ideal ways for robots to help us deal with dynamic, uncertain, and manipulative contexts: by shifting our behavior in situ or helping us build up relevant competencies.

---

[38] Kozyreva, A., Lewandowsky, S. and Hertwig, R., Citizens versus the Internet: Confronting Digital Challenges with Cognitive Tools. Psychological Science in the Public Interest 21(3), 103–156, 2020. https://doi.org/10.1177/1529100620946707.

[39] Thaler, R. and Sunstein, C., Nudge: The Gentle Power of Choice Architecture. New Haven, CT: Yale, 2008.

Robots that boost our abilities and nudge behavior are nascent. Borenstein and Arkin,[42] for instance, imagined the scenario of a robot that helps us build greater empathy among ourselves, human to human. They raise various ethical issues about this. Can we trust robots to act when they need to, in a way that is empathy affirming and not empathy averting? Mehenni et al.[43] deployed Pepper to play a dictator game with nearly 200 French school children. They found that Pepper was more successful than a human facilitator at nudging the children. This raises the question of trust, and perhaps overtrust, by younger folk on robots. Skews, Amodio, and Seibt[44] proposed the idea of fair proxy communication, whereby a telepresence robot intervenes in human–human interactions to correct biased decision-making and ensure that the relationship between the people involved is equitable. They call this a form of "ethical nudging." To explore this idea, they proposed the use of a socially ambiguous robot, such as the Telenoid™ R1 robot, as a mediator between people communicating under controlled conditions. The robot could either obscure social identity characteristics or replace those characteristics, thus averting biased responses from the other party. All of these possibilities are in the early stages but have the potential to change our behavior and help us learn how to do better when in situations saturated with bias.

To this end, we can engage with the following topics on where robots may be able to step in when we are tripped up by our biases:

**Critical next steps:**

- Building on the literature above, we need user studies and human subjects experiments focusing on perceived acceptability, perceived trust, and performance at the intervention task.

**Social factors:**

- We must recognize that robots may need to be trained on socially sensitive material and consider how this may affect participants and eventual users.
- Social cues, whether intentionally designed into the robot or ascribed to the robot by users, may influence the robot's effectiveness or trigger stereotype threats, thus creating a situation of diminishing returns. Longitudinal work can explore this.

**Legal factors:**

- If nudging goes wrong and boosting fails, leading to damages, the courts may need to determine who is responsible – the robot, the creators, or another party.

**Ethical factors:**

- Nudges could nudge the wrong way if the robot fails to act when it should, incidentally encouraging negative behavior; this may be especially pronounced for younger people and vulnerable populations, who may not have the ability to judge or even overtrust false nudges from a robot.
- We need to be aware that obscuring social characteristics to avoid biased reactions could end up deepening existing biases based on social expectations; that is, an assumption that the other party is male because male is the default.

### 19.2.2.3   *Example: Robots That Capitalize on Our Biases … for Our Own Good*

The experience of biases may or may not be good for us or others. Perhaps a robot could be designed to harness our propensity to be biased … in a way beneficial to us. Obo et al.[45] developed an exercise trainer robot for older adults who need motivation to be more physically fit. They made use of the framing effect bias, whereby the framing, positive or negative, of the choices we are given influences our ultimate decision. To do this, the researchers modulated the voice of a semihumanoid PALRO robot in more or less positive or negative ways. While the study sample was small, consisting of seven older adults, the intervention was conducted over several days. In the first four days, the robot made use of the framing effect. Results indicated greater performance when a positive framing was applied. Moreover, when the older adults did poorly, they tended to blame the robot, and when they did well, they attributed that success to themselves; this is known as the self-serving bias. In short, people can be biased to exercise better when a robot employs a positive framing of the exercise, and the worst that can happen is that the robot is blamed for the person's own shortcomings.

---

45 Obo, T., Kasuya, C., Sun, S., and Kubota, N., Human-robot interaction based on cognitive bias to increase motivation for daily exercise. In 2017 IEEE International Conference on Systems, Man, and Cybernetics (SMC), pp. 2945–2950, October 2017. IEEE. https://doi.org/10.1109/SMC.2017.8123075.

In a different context, Johansen, Jensen, and Bemman[46] crafted a mechanical, almost object-shaped robot named "O" that made use of the Dunning–Kruger effect, whereby people overestimate their own abilities and underestimate the abilities of others. Deployed at an art show and discussion setting, the robot carried out an interactive, creative drawing task with passersby. In total, thirty-six people joined the biased group and forty interacted with the unbiased version of the robot. Follow-up interviews indicated that the biased robot was viewed far less favorably than the unbiased robot, even while the biased robot elicited more creative drawings that deviated from the template compared to the unbiased robot. If the goal is creativity, a biased robot may help us achieve it, but at the loss of good feelings for that robot. We may need to consider trade-offs like these and perhaps find ways that they can be counteracted without sacrificing the main effect.

Robots may be able act on their knowledge of our biases in ways that benefit us. While this literature presents a starting point, there is much work to be done, including but not limited to:

**Critical next steps:**

We must identify what biases may elicit benefits for us even while these biases take advantage of "weaknesses" in our cognition. We can then design robots that react to these biases in beneficial ways for us.

**Social factors:**

- We must consider that biased robots may be more effective but less socially acceptable, especially if we recognize that these biases are being harnessed.

---

[46] Johansen, J. J., Jensen, L. G., and Bemman, B., Evaluating Interactions with a Cognitively Biased Robot in a Creative Collaborative Task. In Interactivity, Game Creation, Design, Learning, and Innovation. Cham: Springer, 138–157, 2019. https://doi.org/10.1007/978-3-030-53294-9_10.

- We may need to construct and explore the efficacy of measures to counteract the negative effects that this may have on social acceptability.

**Legal factors:**

- People may have an adverse reaction to being tricked or "scammed." We may need special ethics or even regulations in place to account for negative reactions, so as to protect all parties involved.

**Ethical factors:**

- People may not agree with being taken advantage of, even if the outcome is beneficial. Participation will need to be voluntary.

### 19.2.3   CASE: TRUST

Trust is a fundamental mediating factor of relationships between people, other animals, and artificial agents, too. Trust is a multidimensional concept.47 Many researchers within HRI refer to the definition offered by Wagner and colleagues,48 where trust is "a belief, held by the trustor that the trustee will act in a manner that mitigates the trustor's risk in a situation in which the trustor has put its outcomes at risk."49 When it comes to robots, this means that a person trusts the robot to act accordingly in situations where the person is at risk. Risk can be physical, emotional, psychological, economic ... any situation in which the person has a stake. Trust can be calibrated by the actual capabilities of the robot50 leading to calibrated trust, meaning that there is a match between the person's belief and actual abilities of the robot, distrust, which we may also call undertrust, and overtrust. Ideally, we achieve a Goldilocks level of trust with a robot and maintain this state over time. But various factors intervene to make achieving this equilibrium of trust difficult. It may be that the relative human-likeness of a robot's appearance, especially its uncanniness, could influence feelings of trust toward it. I consider the more typical states of overtrust and undertrust or distrust, with an eye to helping people avoid over- or undertrusting robots as well as making use of our trust biases in different situations.

---

47 Barber, B., The Logic and Limits of Trust. New Brunswick, NJ: Rutgers University Press, 1983; Rempel, J. K., Holmes, J. G., and Zanna, M. P., Trust in Close Relationships. Journal of Personality and Social Psychology 49(1), 95, 1985; Luhmann, N., Trust and Power, Hoboken, NJ: John Wiley & Sons, 2018; Ullrich, D., Butz, A., and Diefenbach, S., The Development of Overtrust: An Empirical Simulation and Psychological Analysis in the Context of Human–Robot Interaction. Frontiers in Robotics and AI, 8, 554578, 2021. https://doi.org/10.3389/frobt.2021.554578; Ueno, T., Sawa, Y., Kim, Y., Urakami, J., Oura, H., and Seaborn, K., Trust in Human-AI Interaction: Scoping Out Models, Measures, and Methods. In CHI Conference on Human Factors in Computing Systems Extended Abstracts, pp. 1–7, April 2022. https://doi.org/10.1145/3491101.3519772.

48 Wagner, A. R., The Role of Trust and Relationships in Human-Robot Social Interaction. Georgia Institute of Technology, 2009; Wagner, A. R., Robinette, P., and Howard, A., Modeling the Human-Robot Trust Phenomenon: A Conceptual Framework based on Risk. ACM Transactions on Interactive Intelligent Systems (TiiS) 8(4), 1–24, 2018. https://doi.org/10.1145/3152890.

49 Wagner (2009, p. 31), supra note 48, at 13.

50 Muir, B. M., Trust between Humans and Machines, and the Design of Decision Aids. International Journal of Man-Machine Studies 27(5–6), 527–539, 1987. https://doi.org/10.1016/S0020-7373(87)80013-5; Lee, J. D. and See, K.

#### 19.2.3.1   Example: When We Trust Robots Too Much

As robots and other intelligent systems have entered daily life in greater and greater numbers, research has indicated that we may trust these systems too much. In other words, we may overtrust.51 Overtrust can occur due to automation bias, where our beliefs about a system's capabilities are far greater than reality,52 and/or complacency, where we become less attentive to the performance of a system over time.53 Work on overtrust in robots remains nascent, with most work derived from automation research.

Overtrust tends to occur in three general contexts: decision-making, emergencies, and security. Salem et al.54 investigated how people would react to an "erratic" robot that made strange and potentially dangerous requests, including those that would lead to property damage. Alarmingly, participants tended to comply with the requests, even though they later

---

A., Trust in Automation: Designing for Appropriate Reliance. Human Factors 46(1), 50–80, 2004. https://doi.org/10.1518/hfes.46.1.50_30392.

51 Wagner, A. R., Borenstein, J., and Howard, A., Overtrust in the Robotic Age. Communications of the ACM 61(9), 22–24, 2018. https://doi.org/10.1145/3241365.

52 Mosier, K. L., Dunbar, M., McDonnell, L., Skitka, L. J., Burdick, M., and Rosenblatt, B., Automation Bias and Errors: Are Teams Better than Individuals? In Proceedings of the Human Factors and Ergonomics Society Annual Meeting (Vol. 42, No. 3). Los Angeles, CA: SAGE Publications, pp. 201–205, October 1998. https://doi.org/10.1177/154193129804200304.

53 Itoh, M., Toward Overtrust-Free Advanced Driver Assistance Systems. Cognition, Technology & Work 14(1), 51–60, 2012. https://doi.org/10.1007/s10111-011-0195-2.

54 Salem, M., Lakatos, G., Amirabdollahian, F., and Dautenhahn, K., Would You Trust a (Faulty) Robot? Effects of Error, Task Type and Personality on Human-Robot Cooperation and Trust. In 2015 10th ACM/IEEE International Conference on Human-Robot Interaction (HRI), pp. 1–8, March 2015, IEEE. https://doi.org/10.1145/2696454.2696497.

rated their perceptions of its trustworthiness negatively. Gaudiello et al.55 evaluated the extent to which people complied with a robot's suggestions. They found that people tended to change their response in line with what the robot suggested, in other words, a conformity effect. Robinette, Howard, and Wagner56 considered how people might react to an emergency assistive agent during a realistic crisis. They observed whether or not people would follow the directions of a robot during a fire … under varying performance conditions, where the robot made an escalating number of mistakes. To increase ecological validity, the researchers modified the environment with fake smoke and fire alarms. Ninety-five percent of participants followed the robot, even when the robot failed, admitted to making mistakes, or when they were told by facilitators that the robot was broken. In a lower stakes situation, Booth et al.57 investigated dorm safety in a social context. The robot attempted to "piggyback" its way into a dorm with other people at the entry. While 40 percent of individuals allowed it entry, 70 percent of group members allowed the robot to enter, pointing to group effects and potentially peer pressure. This increased to 80 percent when the robot was carrying food; that is, it appeared to be a delivery robot. We appear to overtrust social, humanoid robots regardless of the severity of the situation or risk involved.

I offer an overview of critical next steps and factors below, but for greater depth than I can provide here, I point the reader to the article by Aroyo et al.58 on overtrust.

**Critical next steps:**

- We can explore how the relative anthropomorphism of the robot modulates trust, in both directions (over- and undertrust).
- We can explore how expertise, education, experience, and competencies influence trust in robots, especially when people are technically savvy but underexperienced with robots.

---

55 Gaudiello, I., Zibetti, E., Lefort, S., Chetouani, M., and Ivaldi, S., Trust as Indicator of Robot Functional and Social Acceptance. An Experimental Study on User Conformation to iCub Answers. Computers in Human Behavior 61, 633–655, 2016. https://doi.org/10.1016/j.chb.2016.03.057.

56 Robinette, P., Howard, A., and Wagner, A. R., Conceptualizing Overtrust in Robots: Why Do People Trust a Robot that Previously Failed? In Autonomy and Artificial Intelligence: A Threat or Savior? Cham: Springer, 129–155, 2017. https://doi.org/10.1007/978-3-319-59719-5_6.

57 Booth, S., Tompkin, J., Pfister, H., Waldo, J., Gajos, K., and Nagpal, R., Piggybacking Robots: Human-robot Overtrust in University Dormitory Security. In Proceedings of the 2017 ACM/IEEE International Conference on Human-Robot Interaction, pp. 426–434, March 2017. https://doi.org/10.1145/2909824.3020211.

58 Aroyo, A. M., De Bruyne, J., Dheu, O., Fosch-Villaronga, E., Gudkov, A., Hoch, H., ... and Tamò-Larrieux, A., Overtrusting Robots: Setting a Research Agenda to Mitigate Overtrust in Automation. Paladyn, Journal of Behavioral Robotics 12(1), 423–436, 2021. https://doi.org/10.1515/pjbr-2021-0029.

- We can explore how transparency about a robot's abilities affects overtrust and undertrust – perhaps in unexpected ways. Naïve users may, for instance, have high expectations of robots and be let down when they discover that a given robot cannot meet those expectations. This is sure to modulate trust, but it is hard to say in what direction, and may be dependent on individual factors, such as appreciation for transparency.
- We can explore how our tendency to comply and overtrust robots can be deployed as a strategy in prosocial and educational contexts. Perhaps a robotic teacher that has important knowledge to impart will perform better when overtrusted.

**Social factors:**

- We need to consider whether and how social biases and group effects may increase instances of overtrust with robots outside of the contexts explored above.
- While robot anthropomorphism may lead to overtrust, reduced anthropomorphism may lead to deficits on other measures, including satisfaction and perceptions of friendliness. We need to explore this in controlled experiments.

**Legal factors:**

- We should consider overtrust and undertrust as legal constructs for liability regulations. For example, who is at fault if a driver undertrusts a robotic vehicle and takes control, which leads to an accident that would have been avoided if the robot remained in control?
- Interventions in the design of robots to avoid overtrust may need to consider legal age and disability regulations as well as ethical matters generally. Young children, for instance, may be susceptible but unable to understand an explanation of a robot's true abilities, and thus still overtrust the robot.

**Ethical factors:**

- Certain people may have a propensity to overtrust, such as those naive to robots and artificial intelligence, very young people, and people with dependent personality disorder. These people should be designed for and ideally with to craft HRI scenarios that meet their needs and desires.

### 19.2.4   CASE: STEREOTYPES

Stereotypes are widely held but simple models of people in societies. We rely on stereotypes when encountering and making sense of individuals for the first time. Stereotype activation occurs when we react to models of the people associated with the social group or certain social characteristics that we perceive in an individual. We can react to stereotypes we perceive in others as well as in ourselves. Typically, these models are negative, narrow, simplistic, and yet firmly held, leading to a phenomenon known as stereotype threat.59 When people are reminded, purposefully or not, about social characteristics related to stereotypes in a given situation, they tend to more easily react in line with those stereotypes, a phenomenon called stereotype priming.60 A wealth of research on computer agents and more recently on social robots has suggested that stereotypes are activated even when the merest hint of a cue exists.61 Furthermore, people are mostly unaware that they are doing it. The implications are at least twofold: We can design in stereotypes to elicit certain reactions, or we can avoid stereotyped design choices to evade or provoke certain reactions.

### 19.2.4.1   Example: Robots That Dodge or Disrupt Stereotypes

If we know when stereotypes may be activities, can we avoid stereotype priming and threats? Can we even go so far as to make use of our knowledge of stereotypes to purposefully disrupt expectations, provoke, and inspire reflection or creativity? Scant research on robots exists in this area, but recent work is heading in this direction. Ogunyale, Bryant, and Howard62 considered race stereotypes of emotional expression within the US context ... applied to humanoid robots. Respondents were presented with videos of either a white or black version of the ROBOTIS Darwin-Mini robot performing emotional gestures. Notably, they were not primed with stereotypes related to light-skinned or dark-skinned people and emotional expression. The research design was limited by lack of a direct comparison or manipulation check of the robot's perceived racial alignment. Nevertheless, compared to previous research that did use stereotype priming, there were no significant effects. Spatola et al.[63] presented respondents with images of robots said to have come from certain countries. Different participants were told that the same robot was from different countries. They found that stereotypes about each nation's warmth and competence biased respondents in their perceptions of each robot's warmth and competence, pointing to a stereotype priming effect. While not carried out in this research, such a finding suggests that mere attribution of a national identity to a robot can elicit certain impressions ... a quick and resource-light strategy to influence initial attitudes toward and potentially behavior with a robot.

---

59 Steele, C. M., A Threat in the Air: How Stereotypes Shape Intellectual Identity and Performance. American Psychologist 52(6), 613. https://psycnet.apa.org/doi/10.1037/0003-066X.52.6.613; Spencer, S. J., Steele, C. M., and Quinn, D. M., Stereotype Threat and Women's Math Performance. Journal of Experimental Social Psychology 35(1), 4–28, 1999. https://doi.org/10.1006/jesp.1998.1373; Aronson, J., Lustina, M. J., Good, C., Keough, K., Steele, C. M., and Brown, J., When White Men Can't Do Math: Necessary and Sufficient Factors in Stereotype Threat. Journal of Experimental Social Psychology 35(1), 29–46. https://doi.org/10.1006/jesp.1998.1371; Spencer, S. J., Logel, C., and Davies, P. G., Stereotype Threat. Annual Review of Psychology 67(1), 415–437, 2016. https://doi.org/10.1146/annurev-psych-073115-103235.

60 Shih, M., Pittinsky, T. L., and Ambady, N., Stereotype Susceptibility: Identity Salience and Shifts in Quantitative Performance. Psychological Science 10(1), 80–83, 1999. https://doi.org/10.1111/1467-9280.00111; Steele, J. R. and Ambady, N., "Math is Hard!" The Effect of Gender Priming on Women's Attitudes. Journal of Experimental Social Psychology 42(4), 428–436, 2006. https://doi.org/10.1016/j.jesp.2005.06.003; Gibson, C. E., Losee, J., and Vitiello, C., A Replication Attempt of Stereotype Susceptibility (Shih, Pittinsky, & Ambady, 1999): Identity Salience and Shifts in Quantitative Performance. Social Psychology 45(3), 194, 2014. https://doi.org/10.1027/1864-9335/a000184.

61 Nass, C., Steuer, J., and Tauber, E. R., Computers Are Social Actors. In Proceedings of the SIGCHI Conference on Human Factors in Computing Systems, pp. 72–78, April 1994. https://doi.org/10.1145/191666.191703; Tay, B., Jung, Y., and Park, T., When Stereotypes Meet Robots: The Double-edge Sword of Robot Gender and Personality in Human–robot Interaction. Computers in Human Behavior, 38, 75–84, 2014. https://doi.org/10.1016/j.chb.2014.05.014.

62 Ogunyale, T., Bryant, D. A., and Howard, A., Does Removing Stereotype Priming Remove Bias? A Pilot Human-robot Interaction Study. Presented at the 5th Workshop on Fairness, Accountability, and Transparency in Machine Learning (FAccT 2018), Stockholm, Sweden, July 15, 2018. https://doi.org/10.48550/arXiv.1807.00948.

Song-Nichols and Young[64] carried out a counter-stereotyping experiment with young children about gender roles using robots. The gender stereotype attitudes of children in the counter-stereotyping condition were reduced while those in the stereotyped condition were increased. This indicates that robots can dampen or inflate views of gender at a young age. Winkle et al.[65] proposed three ways in which robots could be deployed to tackle biases and stereotypes related to gender. Specifically, robots designed to encourage girls to engage in engineering and computer science, antisexist robots that confront abusive language and behavior head-on, and robots with feminine gender cues that disrupt expectations about gender norms in their deployment and behavior. Pennefather and I[66] offered another perspective: neutralizing gender. Our rapid review and ongoing living review work critically examines whether and how gender neutrality in robots can be achieved. At present, little can be empirically concluded due to limitations in research designs, assumptions that gender neutrality was established, and a sheer lack of available literature. Nevertheless, early results point to a few promising strategies: reducing the presence of humanlike cues, mixing

---

masculine and feminine gender cues, using gender-neutral names, and most provocatively, shifting the robot's gender over time in a gender fluid manner. Stereotype activation may not always be avoidable, but it seems that it can often be disrupted, if we are aware of the possibility and harness it.

Many opportunities in the area of stereotypes and robotic designs exist, ripe for research and exploration. I offer a few possibilities for consideration:

**Critical next steps:**

- We can explore stereotype activation, stereotype priming, and counter-stereotyping with a variety of robots boosting an array of social characteristics beyond gender and race.
- We can explore how stereotypes intersect to change, augment, or disrupt reactions; that is, intersectionality in stereotypes.
- We can explore whether and how stereotyping can be evaded through the design of the robot, the design of the research, or potentially other means. Are gender-neutral robots possible or simply an ideal, for instance?
- We can explore whether stereotype-disruptive robots also disrupt stereotyped attitudes toward people. For instance, experience with an antistereotype robot could be embedded in research on attitude change toward people.

**Social factors:**

- Stereotypes tend to do more harm than good, so we should avoid reproducing stereotypes in robots for the good of humanity.
- Stereotype-busting robots could shift attitudes about people, but this needs to be studied empirically.

**Legal factors:**

- Nations and states may regulate or even criminalize expressions of hate, discrimination, and so on, and this may extend to robots as a mode of expression such notions in the future.

**Ethical factors:**

- Robots could pave the way for a new ethics of identity. Virtual influencers have paved the way for such phenomena – robots of the near future could do the same.
- Robots are nonetheless artificial and may not be taken seriously as purveyors or disruptors of stereotypes, potentially curdling a backlash.

## 19.3  Track: Against Bias in Robots

Robots may be able to help us manage, prevent, and even reverse biases in decision-making and interacting with the world and each other. But we also need to consider the other side of the coin. As engineers and deployers of robots, our thinking and actions are formalized through research and development practices. In other words, robots can be the bearers of our own biases. The good news is that we need only to turn a critical eye on the fruits of our labors to smoke it out. A recent critical review[67] identified where and how cognitive biases have been found, if not fruitfully deployed, in social robotics work. Yet, most of the work so far has focused on stereotype effects. Early work by Nass, Reeves, Moon, and colleagues[68] revealed pervasive biases in how humanoid computer agents tend to be interpreted by people. These stereotyped responses by people to social constructs embedded in the voice and body of computer agents have been demonstrated time and again over the years, leading to the formulation of a Computer Are Social Actors (CASA) model. Gender,69 race,70 nationality,71 accent,72 age,73 and more are social constructs embedded in the design of robots that are known to prompt biased reactions in participants. In light of technological advances and social movements, a new wave of efforts has pushed the issue of bias to the next level. Visual processing systems used for object detection and facial recognition have been found to be biased in gender, race, and age.74 We must recognize this as a fundamentally human problem. Technology is not neutral, however advanced or formalized or common it may be. We must make concerted efforts against embedding our biases into robots and designing robots in biased ways, at least unawares and without good reasons to do so. I now turn to common and crucial ways in which bias has seeped into the design and deployment of robots, and what we can do about it.

---

67 Letheren, K., Russell-Bennett, R., Whittaker, L., Whyte, S., and Dulleck, U., The Evolution Is Now: Service Robots, Behavioral Bias and Emotions. Emotions and Service in the Digital Age, 2020. https://doi.org/10.1108/S1746-979120200000016005.

68 Nass, Steuer, and Tauber (1994), supra note 61, at 16; Nass, C., Moon, Y., Fogg, B. J., Reeves, B., and Dryer, D. C., Can Computer Personalities be Human Personalities? International Journal of Human-Computer Studies 43(2), 223–239, 1995. https://doi.org/10.1006/ijhc.1995.1042; Reeves, B. and Nass, C., The Media Equation: How People Treat Computers, Television, and New Media Like Real People. Cambridge, UK: Cambridge University, 1996; Nass, C. I., Moon, Y., and Morkes, J. Computers are Social Actors: A Review of Current Research. Human Values and the Design of Computer Technology 72, 137, 1997; Nass, C. and Moon, Y., Machines and Mindlessness: Social Responses to Computers. Journal of Social Issues 56(1), 81–103, 2000. https://doi.org/10.1111/0022-4537.00153.

---

69 Eyssel, F. and Hegel, F., (S) he's Got the Look: Gender Stereotyping of Robots 1. Journal of Applied Social Psychology 42(9), 2213–2230, 2012. https://doi.org/10.1111/j.1559-1816.2012.00937.x; Alesich, S. and Rigby, M., Gendered Robots: Implications for our Humanoid Future. IEEE Technology and Society Magazine 36(2), 50–59, 2017. https://doi.org/10.1109/MTS.2017.2696598; Okanda, M. and Taniguchi, K., Is a Robot a Boy? Japanese Children's and Adults' Gender-Attribute Bias toward Robots and Its Implications for Education on Gender Stereotypes. Cognitive Development 58, 101044, 2021. https://doi.org/10.1016/j.cogdev.2021.101044; Wang, Z., Huang, J., and Fiammetta, C., Analysis of Gender Stereotypes for the Design of Service Robots: Case Study on the Chinese Catering Market. In Designing Interactive Systems Conference 2021, pp. 1336–1344, June 2021. https://doi.org/10.1145/3461778.3462087; Winkle et al. (2021), supra note 65, at 17; Perugia et al. (2022), supra note 12, at 3; Suzuki, T. and Nomura, T., Gender Preferences for Robots and Gender Equality Orientation in Communication Situations. AI & SOCIETY, 1–10, 2022. https://doi.org/10.1007/s00146-022-01438-7.

70 Bartneck, C., Yogeeswaran, K., Ser, Q. M., Woodward, G., Sparrow, R., Wang, S., and Eyssel, F., Robots and Racism. In Proceedings of the 2018 ACM/IEEE International Conference on Human-Robot Interaction, pp. 196–204, February 2018. https://doi.org/10.1145/3171221.3171260; Addison, A., Bartneck, C., and Yogeeswaran, K., Robots Can be More than Black and White: Examining Racial Bias towards Robots. In Proceedings of the 2019 AAAI/ACM Conference on AI, Ethics, and Society, pp. 493–498, January 2019. https://doi.org/10.1145/3306618.3314272.

71 Spatola et al. (2019), supra note 63, at 16.

72 Torre, I. and Le Maguer, S., Should Robots have Accents? In 2020 29th IEEE International Conference on Robot and Human Interactive Communication (RO-MAN), pp. 208–214, August 2020. IEEE. https://doi.org/10.1109/RO-MAN47096.2020.9223599.

73 Perugia et al. (2022), supra note 12, at 3.

74 Howard, A. and Kennedy III, M., Robots Are Not Immune to Bias and Injustice. Science Robotics 5(48), eabf1364, 2020. https://doi.org/10.1126/scirobotics.abf1364.

## 19.3.1  CASE: INTERSECTIONAL DESIGN

Intersectionality describes how social characteristics intersect with power in compounding ways. Originally proposed by Crenshaw[75] for law, intersectionality has been taken up as a broader explanatory framework … one that is critical for robots, as well. Designing and deploying robots is a matter of power. Engineers, roboticists, designers, developers, programmers, technicians, practitioners, researchers, doctors, nurses, and other stakeholders on "our" side are in a position to make decisions or unwittingly embed our own perspectives, choices, and demands into robots. When we create humanoid robots, do we consider how the embodiment of these robots invokes gender, age, ethnicity/race, class, and other social characteristics? Do we consider how these characteristics can intersect, or do we only consider one at a time? Do we recognize that others, including if not especially our intended end users, may have a different perspective on how we have chosen to embed these characteristics in the robot's form factor and morphology, verbal and nonverbal expressions, role and application, machine learning algorithms, and underlying AI systems? Critical perspectives within HRI and related spaces have raised awareness of how our own unconscious biases can play a role in representation and social harm through how we design and use robots with people.[76] Intersectional design[77] has been proposed as a framework relevant to the design of socially expressive robots that have human characteristics in appearance and behavior. Work is just starting, and there is a lot of ground to cover. I consider three key elements in robots that engage with people: morphology and form factor, including "voice" and "body"; algorithms and data sets that make up the "brain"; and behaviors or "actions," both passive and reactive.

---

### 19.3.1.1   *Example: Embodying Diversity in Voice and Body*

Social robots have a physical "body" or form factor. The morphology – shape, size, width, weight, height, materials, textures, "clothes," face, limbs, wheels, and so on – are taken in and processed by our minds in certain ways. Moreover, the voice of the robot can influence perceptions,78 even to the point of causing confusion and uncanny valley reactions when gender cues do not align.79 When humanlike cues exist, we tend to draw on human models to understand what we are perceiving and how to react to it – and mostly this is a rapid and unconscious process. Indeed, while we may make decisions about what a robot should look like, we may not always do so with all possibilities in mind. Similarly, people interacting with robots tend to interpret those robots in line with human models, including any limited, stereotyped, or otherwise negative associations they may have about people when it comes to the specific social characteristics those robots embody. For example, while we may take it for granted that a robot is gender neutral, our participants may not agree.80 Robots, especially socially expressive ones that feature humanoid markers, are not tabula rasa.

The social robotics community has started to take notice. As recent workshops on critical studies for HRI have indicated, there is a pressing need and desire to eliminate social biases in the design of robots and their deployment in real life and research.[81] Social characteristics that trace back to models of people have been recognized, as outlined above. Yet, the path forward is a bit unclear. For instance, we may racialize a relatively "light" or "dark" "skinned" robot and react in line with human stereotypes. Bartneck et al.[82] replicated a famous study by Eberhardt et al.[83] on race and racism within the American context: Implicit anti-Black biases triggered by images of guns. They found that this anti-Black association held true for both images of Black people and robots racialized as Black by adjusting the "skin" color of the robot. In contrast, Bryant, Borenstein, and Howard[84] explored whether and how gendered robots affect trust under different occupational settings. Against a backdrop of previous research and gender stereotyped assumptions, they found that occupational performance predicted levels of trust rather than gender. Perugia et al.[85] present another nuanced perspective, considering people's social categorization of a large range of robot images in terms of gender and age. They found that attributions of masculinity tended to align with body shape, while attributions of femininity were associated with superficial features, such as eyelashes. Age wise, most robots were perceived as adults, and age was associated with lack of facial features, for reasons that are difficult to tease out.

---

78 Walters, M. L., Syrdal, D. S., Koay, K. L., Dautenhahn, K., and Te Boekhorst, R., Human Approach Distances to a Mechanical-Looking Robot with Different Robot Voice Styles. In RO-MAN 2008-The 17th IEEE International Symposium on Robot and Human Interactive Communication, pp. 707–712, August 2008. IEEE. https://doi.org/10.1109/ROMAN.2008.4600750; Crowelly, C. R., Villanoy, M., Scheutzz, M., and Schermerhornz, P., Gendered Voice and Robot Entities: Perceptions and Reactions of Male and Female Subjects. In 2009 IEEE/RSJ International Conference on Intelligent Robots and Systems, pp. 3735–3741). October 2009. IEEE. https://doi.org/10.1109/IROS.2009.5354204; Eyssel, F., Kuchenbrandt, D., Bobinger, S., De Ruiter, L., and Hegel, F., "If You Sound Like Me, You Must Be More Human' on the Interplay of Robot and User Features on Human-Robot Acceptance and Anthropomorphism. In Proceedings of the Seventh Annual ACM/IEEE International Conference on Human-Robot Interaction (HRI), pp. 125–126, March 2012. https://doi.org/10.1145/2157689.2157717; McGinn, C. and Torre, I., Can You Tell the Robot by the Voice? An Exploratory Study on the Role of Voice in the Perception of Robots. In 2019 14th ACM/IEEE International Conference on Human-Robot Interaction (HRI), pp. 211–221, March 2019. IEEE. https://doi.org/10.1109/HRI.2019.8673305.

79 Takayama, L., Groom, V., and Nass, C., I'm Sorry, Dave: I'm Afraid I Won't Do that: Social Aspects of Human-agent Conflict. In Proceedings of the SIGCHI Conference on Human Factors in Computing Systems, pp. 2099–2108, April 2009. https://doi.org/10.1145/1518701.1519021; Mitchell, W. J., Szerszen Sr, K. A., Lu, A. S., Schermerhorn, P. W., Scheutz, M., and MacDorman, K. F., A Mismatch in the Human Realism of Face and Voice Produces an Uncanny Valley. i-Perception 2(1), 10–12. https://doi.org/10.1068/i0415.

80 Seaborn and Frank (2022), supra note 6, at 3.

This body of work asks us to question where these reactions are coming from: Our set models of the (human) world, or cues embedded in social robots, knowingly or unknowingly by their creators? Likely, it is a combination of the two. However, we only have so much control over the reaction of other people to our robotic stimuli. Still, we can be conscious about how we design robots and how others are likely to react to those designs. For instance, children, including toddlers, are already socialized with ideas of gender[86] and race,[87] and apply these models to robots. Yet, interventions may be staged or embedded in the design of robots. We can draw inspiration from related domains. For example, Powell-Hopson and Hopson[88] demonstrated that an intervention could shift children's preferences toward dolls racialized as white or black, severely reducing a preexisting white bias. Social robots likewise may be designed to disrupt and shift expectations through bold decisions in their appearance that aim for diversity, antistereotyping, and prosocial epiphanies.

---

81 Lee, H. R., Cheon, E., De Graaf, M., Alves-Oliveira, P., Zaga, C., and Young, J., Robots for Social Good: Exploring Critical Design for HRI. In 2019 14th ACM/IEEE International Conference on Human-Robot Interaction (HRI), pp. 681–682, March 2019. IEEE. https://doi.org/10.1109/HRI.2019.8673130.

82 Bartneck et al. (2018), supra note 70, at 19.

83 Eberhardt, J. L., Goff, P. A., Purdie, V. J., and Davies, P. G., Seeing Black: Race, Crime, and Visual Processing. Journal of Personality and Social Psychology 87(6), 876, 2004. https://psycnet.apa.org/doi/10.1037/0022-3514.87.6.876.

84 Bryant, D. A., Borenstein, J., and Howard, A., Why Should We Gender? The Effect of Robot Gendering and Occupational Stereotypes on Human Trust and Perceived Competency. In Proceedings of the 2020 ACM/IEEE International Conference on Human-Robot Interaction, pp. 13–21, March 2020. https://doi.org/10.1145/3319502.3374778.

85 Perugia et al. (2022), supra note 12, at 3.

To explore how to achieve diversity in robot voice and body and what effects diversity in robot voice and body may have, we can consider the following ideas:

**Critical next steps:**

- We can conduct systematic surveys on whether and how specific social characteristics trigger stereotyped and negative reactions.
- We can carry out empirical work on neutrality in robots: gender, race, accent, and so on.
- We can explore whether young children have a race, ethnicity, language, accent, and/or age bias toward robots. If successful, robots may then be used as a more ethical stimulus for assessing what biases people have and from what age, even the youngest and most vulnerable among us.

**Social factors:**

- We must be aware of how we design in social cues drawn from human models, as we may unintentionally reinforce negative associations and replicate stereotypes into the design of robots.
- We may also unintentionally rely on a limited vision of what robots look, sound, and act like. Conscious efforts toward diversity will help mitigate our own biases.
- Robots may be used to challenge stereotypes an disrupt expectations. Robots could be creatively designed to model new visions of human identities.

---

86 Okanda & Taniguchi (2021), supra note 69, at 19.

87 Bartneck et al. (2018), supra note 70, at 19.

88 Powell-Hopson, D. and Hopson, D. S., Implications of Doll Color Preferences among Black Preschool Children and White Preschool Children. Journal of Black Psychology 14(2), 57–63, 1988. https://doi.org/10.1177/00957984880142004.

**Legal factors:**

- Robots designed in stereotyped ways that reflect negatively on real people could constitute a form of discrimination or even hate crime.

**Ethical factors:**

- People deserve access to a variety of robots that are like and unlike them. Customizable robots, like customizable avatars in games and apps, could be a novel diversity measure embedded in future robot bodies.

*19.3.1.2   Example: From Algorithmic Bias to Diverse Data Sets and Explainable Intelligence*

Algorithmic bias occurs when the functions in an algorithm, its approach to prediction, and/or its training data are biased. This is not a machine problem; it is a human problem. People are the ones creating and making decisions about machines. Part of the problem is that people are not aware of their own biases. Another part of the problem is that we do not have access to the heart of the algorithms, either due to proprietary restrictions or their black-box nature. These algorithms are now finding a home in the brains of social robots. Aldebaran-SoftBank's humanoid Pepper, for instance, uses a proprietary set of algorithms to detect faces and facial expressions.[89] While not robots per se, we may think of algorithms as comprising the robot "brain." As such, these algorithms are becoming an important feature of HRI experiences and the ways in which they can bias or be biased.

Tackling algorithmic bias is challenging, but there are two uncontroversial aspects that we can focus on: data sets and explainability. Data sets are used to train and test the algorithms. They have to be big, but they also have to be accurate. This is where bias can creep in: The data sets being fed to the algorithms represent a limited or disproportionate section of the real data out in the world. In other words, they are not diverse enough. Explainability, commonly captured under the notion of explainable AI or XAI,[90] refers to an algorithm that provides its decision-making processes openly and in a way that a person can understand. Avoiding algorithmic bias involves two feats: Ensuring that the data is sufficiently large and diverse, and providing a means by which to understand why the algorithms have acted on that data in the ways that we can observe. At present, both remain a challenge to detect and enact.

---

[89] This is according to the developers; refer to Pandey, A. K. and Gelin, R., A Mass-Produced Sociable Humanoid Robot: Pepper: The First Machine of Its Kind. IEEE Robotics & Automation Magazine 25(3), 40–48, 2018. https://doi.org/10.1109/MRA.2018.2833157. However, Pepper's real abilities have been questioned; refer to Robertson, J., Robo Sapiens Japanicus: Robots, Gender, Family, and the Japanese Nation, Oakland, CA, USA: University of California Press, 2017. https://doi.org/10.1525/9780520959064.

[90] Gunning, D., Stefik, M., Choi, J., Miller, T., Stumpf, S., and Yang, G. Z., XAI – Explainable Artificial Intelligence. Science Robotics 4(37), eaay7120, 2019. https://doi.org/10.1126/scirobotics.aay7120.

Social robots are no exception. Hundt and colleagues discovered widespread algorithmic bias in a common model used in robots called OpenAI CLIP.91 The researchers used a range of stimuli made up of people's faces – race- and gender-diverse – along with task descriptions that contained stereotyped language. In doing so, they tested two common forms of data and processing for social robots: natural language processing (NLP) and computer vision. Importantly, they did not focus on one social feature alone, considering the intersection of race and gender. This harkens back to Crenshaw's ground-breaking concept of intersectionality.92 This phenomenon on social characteristics and legal structures was revealed in the context of human–human discrimination at the intersection of race and gender, specifically Black women in the American context. Intersectionality is rapidly being taken up as a key approach to understanding algorithmic bias. We cannot always focus on one factor in isolation; instead, we have to consider how a variety of factors interrelate and mediate forms of bias and discrimination. Buolamwini and Gebru's landmark paper93 on computer vision and algorithmic bias considered the intersection of race and gender.94 Later, Bryant and Howard95 found that facial recognition algorithms geared toward detecting emotional expressions were biased against children. On the language front, Bolukbasi et al.96 interrogated gender biases in word embeddings, a popular approach to NLP that is employed across a variety of platforms, from search engines to conversational agents to social robots. In short, the eyes and ears of our robot companions may be biased. This may make them more humanlike, in a way, but that begs the question: What form of humanity do we wish to replicate in our mechanical doppelgangers?

---

91 Hundt, A., Agnew, W., Zeng, V., Kacianka, S., and Gombolay, M., Robots Enact Malignant Stereotypes. In Proceedings of the 2022 ACM Conference on Fairness, Accountability, and Transparency (FAccT '22), pp. 743–756, June 2022. https://doi.org/10.1145/3531146.3533138.

92 Crenshaw, K., Demarginalizing the Intersection of Race and Sex: A Black Feminist Critique of Antidiscrimination Doctrine, Feminist Theory and Antiracist Politics. In Feminist Legal Theory, Oxford, UK: Routledge, 23–51, 2013. https://chicagounbound.uchicago.edu/uclf/vol1989/iss1/8/.

93 Buolamwini, J. and Gebru, T., Gender Shades: Intersectional Accuracy Disparities in Commercial Gender Classification. In Conference on Fairness, Accountability and Transparency, pp. 77–91. PMLR, January 2018. http://proceedings.mlr.press/v81/buolamwini18a.html.

94 Howard, A., Real Talk: Intersectionality and AI. MIT Sloan Management Review, 2021. https://sloanreview.mit.edu/article/real-talk-intersectionality-and-ai/.

95 Bryant, D. A. and Howard, A., A Comparative Analysis of Emotion-Detecting AI Systems with Respect to Algorithm Performance and Dataset Diversity. In Proceedings of the 2019 AAAI/ACM Conference on AI, Ethics, and Society, pp. 377–382, January 2019. https://doi.org/10.1145/3306618.3314284.

96 Bolukbasi, T., Chang, K. W., Zou, J. Y., Saligrama, V., and Kalai, A. T., Man Is to Computer Programmer as Woman Is to Homemaker? Debiasing Word Embeddings. Advances in Neural Information Processing Systems (NIPS 2016), 29, 2016. https://proceedings.neurips.cc/paper/2016/hash/a486cd07e4ac3d270571622f4f316ec5-Abstract.html.

To explore this question, we can strive to improve our data sets, algorithms, and explanations of how these components work together to drive the behavior and decision-making in robots. I offer the following possibilities for exploration:

**Critical next steps:**

- We can empirically investigate and debias commercial and research robots for social biases using an intersectional lens that includes multiple factors and how they intersect.
- We can empirically investigate and debias the data sets and data sources being fed into social robots – what is training their eyes and ears.
- We can form coalitions with industry to access and assess propriety algorithms used in the creation of socially expressive robots, or at least their social expressions.

**Social factors:**

- Algorithms should be developed with human diversity in mind; otherwise, robots will discriminate based on one or more social features.

**Legal factors:**

- Intersectionality has legal roots. Algorithms that bias in multiple and cascading ways may have different and possibly more severe legal repercussions.
- Laws about intersectionality may need to be extended to social robots.

**Ethical factors:**

- Algorithms are for everyone. Every robot (or its "brain") should be able to recognize and treat all people with equity, regardless of their apparent gender, race, age, or any other social characteristic.

### 19.3.1.3   Example: Behavior as a Two-Way Street

Bias can be reflected in the behavior we design into robots. We may assume that everyone will treat robots in line with our own imaginations or expectations, but that is not necessarily the case. People can and do abuse robots, or at least act in abusive ways toward robots.[97] Abusive behavior toward robots in general98 and particularly in relation to social characteristics like gender99 and age, especially children,100 has been a longstanding point of concern in HRI. Concern about abuse toward robots has even led to persuasive arguments for their legal protection.101 We can design robots that fight back against abuse. As Keijsers and Bartneck102 warn, "mindless robots get bullied." This is not the fault of the robot – this is on our end. We need to mindfully and critically recognize that people are not always nice to each other, and potentially not robots, as well. This is not just bad "for robots"; interacting with robots that are abusive or allow abusive behavior can have negative repercussions for us. We must not forget that the greater proportion of HRI work is Wizard of Oz,103 meaning that a human actor is secretly controlling the behavior of the robot behind the scenes. Rea, Geiskkovitch, and Young104 ask us to consider the impact of negative and otherwise emotionally or socially difficult studies for the human Wizard – ethics boards typically do not assess this, after all. In general, we cannot be "abuse evasive" and ignore the issue. For the good of robots and ourselves, we need to design around potential abusive and other toxic reactions to robots.

---

[97] Whitby, B., Sometimes It's Hard to be a Robot: A Call for Action on the Ethics of Abusing Artificial Agents. Interacting with Computers 20(3), 326–333, 2008. https://doi.org/10.1016/j.intcom.2008.02.002.

---

98 Bartneck, C., Rosalia, C., Menges, R., and Deckers, I., Robot Abuse-A limitation of the Media Equation. Presented at Workshop on Abuse (Interact 2005). http://hdl.handle.net/10092/16925; Bartneck, C. and Hu, J., Exploring the Abuse of Robots. Interaction Studies 9(3), 415–433, 2008. https://doi.org/10.1075/is.9.3.04bar; Bartneck, C. and Keijsers, M., The Morality of Abusing a Robot. Paladyn, Journal of Behavioral Robotics 11(1), 271–283, 2020. https://doi.org/10.1515/pjbr-2020-0017

99 Winkle et al. (2021), supra note 65, at 17; Winkle, K., Jackson, R. B., Melsión, G. I., Bršćić, D., Leite, I. and Williams, T., Norm-breaking Responses to Sexist Abuse: A Cross-cultural Human Robot Interaction Study. In Proceedings of the 2022 ACM/IEEE International Conference on Human-Robot Interaction, pp. 120–129, March 2022. https://dl.acm.org/doi/abs/10.5555/3523760.3523780.

100 Bršćić, D., Kidokoro, H., Suehiro, Y., and Kanda, T., Escaping from Children's Abuse of Social Robots. In Proceedings of the 10th Annual ACM/IEEE International Conference on Human-Robot Interaction, pp. 59–66, March 2015. https://doi.org/10.1145/2696454.2696468; Nomura, T., Uratani, T., Kanda, T., Matsumoto, K., Kidokoro, H., Suehiro, Y., and Yamada, S., Why Do Children Abuse Robots? In Proceedings of the Tenth Annual ACM/IEEE International Conference on Human-Robot Interaction Extended Abstracts, pp. 63–64, March 2015. https://doi.org/10.1075/is.17.3.02nom; Ku, H., Choi, J. J., Lee, S., Jang, S., and Do, W., Designing Shelly, a Robot Capable of Assessing and Restraining Children's Robot Abusing Behaviors. In Companion of the 2018 ACM/IEEE International Conference on Human-Robot Interaction, pp. 161–162, March 2018. https://doi.org/10.1145/3173386.3176973; Yamada, S., Kanda, T., and Tomita, K., An Escalating Model of Children's Robot Abuse. In 2020 15th ACM/IEEE International Conference on Human-Robot Interaction (HRI), pp. 191–199, March 2020. IEEE.

101 Darling, K., Extending Legal Protection to Social Robots: The Effects of Anthropomorphism, Empathy, and Violent Behavior towards Robotic Objects. In Robot law. Northampton, MA, USA: Edward Elgar Publishing, 2016. https://doi.org/10.4337/9781783476732.00017.

102 Quote is the title of Keijsers, M. and Bartneck, C., Mindless Robots Get Bullied. In 2018 13th ACM/IEEE International Conference on Human-Robot Interaction (HRI), pp. 205–214, March 2018. IEEE.

103 Riek, L. D., Wizard of oz Studies in HRI: A Systematic Review and New Reporting Guidelines. Journal of Human-Robot Interaction 1(1), 119–136, 2012. https://doi.org/10.5898/JHRI.1.1.Riek.

Researchers have started to explore how robots can be designed to raise awareness of abuse and fight back against it – including by eliciting positive, prosocial responses from people. This can mean one or both of changing the robot's behavior and/or influencing the human's behavior through the design of the robot. Tan et al.105 created a scenario wherein an actor abused a robot in front of a naive participant. The robot, NAO, was designed to respond in one of three ways: ignore it; shut down; or have an emotional reaction. The idea was to explore whether and under what situations would people step up and defend the robot … or fall prey to the bystander effect,106 whereby people expect others to intervene and therefore do nothing, even when they could and would act. Most people had the strongest response when the robot shut down, but no particular reaction from the robot led to an increase or decrease in taking action in the moment. While not good for robot kind, this does reflect patterns in human behavior among humans.108 In a similar study, a robot was abused by an actor in Connolly et al.,107 but in this scenario the response of robot bystanders was manipulated: either an emotional outburst or ignoring the abuse of their robot friend. Participants were more likely to respond when the emotional cues were present. Nevertheless, in both of these studies, participants' existing attitudes of empathy had no relationship to their responses, making it difficult to predict who may take action. Moreover, not all participants reacted in kind.

More research is needed to develop innovative ways in which people can be urged to take action or be interrupted if they resort to negative behavior. To this end, we can consider the following factors in design and research work:

---

104 Rea, D. J., Geiskkovitch, D., and Young, J. E., Wizard of Awwws: Exploring Psychological Impact on the Researchers in Social HRI Experiments. In Proceedings of the Companion of the 2017 ACM/IEEE International Conference on Human-Robot Interaction, pp. 21–29, March 2017. https://doi.org/10.1145/3029798.3034782.

105 Tan, X. Z., Vázquez, M., Carter, E. J., Morales, C. G., and Steinfeld, A., Inducing Bystander Interventions during Robot Abuse with Social Mechanisms. In 2018 13th ACM/IEEE International Conference on Human-Robot Interaction (HRI), pp. 169–177, March 2018. IEEE.

106 Fischer, P., Krueger, J. I., Greitemeyer, T., Vogrincic, C., Kastenmüller, A., Frey, D., … and Kainbacher, M., The Bystander-Effect: A Meta-Analytic Review on Bystander Intervention in Dangerous and Non-dangerous Emergencies. Psychological Bulletin 137(4), 517, 2011. https://psycnet.apa.org/doi/10.1037/a0023304.

107 Connolly, J., Mocz, V., Salomons, N., Valdez, J., Tsoi, N., Scassellati, B., and Vázquez, M., Prompting Prosocial Human Interventions in Response to Robot Mistreatment. In Proceedings of the 2020 ACM/IEEE International Conference on Human-Robot Interaction, pp. 211–220, March 2020. https://doi.org/10.1145/3319502.3374781.

**Critical next steps:**

- We can explore whether and how these results play out in the wild, not just the lab. Ideally, we can try to capture situations where people are at their most natural, feeling comfortable and unobserved.
- We can explore new ways of approaching abuse in interactions with robots. For instance, intent and context may play a role in how best to respond to abuse.
- We can explore robots that detect bullying in children and react with teaching moments.

**Social factors:**

- Avoiding the possibility of people treating robots badly means that the robot's response could negatively reinforce poor treatment.
- Age and social naivety may be particularly important to consider when deciding whether and how to design robots that react to abusive behavior.

**Legal factors:**

- Robots that fight back need to be carefully designed so as not to be abusive themselves. Roboticists and researchers could be liable if a person feels that they have been mistreated by a robot.

**Ethical factors:**

- Robot abuse might not be taken seriously, but even if robots are not "people" or "living beings" or "true intelligence," arguably all things in this world are due respect.
- Robots can help us learn the ethics of treating others with respect.

### 19.3.2   CASE: DIVERSITY IN PRAXIS

"We are better together" could be an alternative title for this section. Reducing and avoiding bias in the design and deployment of robots is surely a team effort, especially because we implicitly bias the robots we design and use.108 For this, team diversity is key.109 Robotics is a gateway to STEM education and careers.110 Robots are cool! But robotics is not a diverse field of study. Indeed, critical voices drawing have highlighted a severe and ongoing diversity problem in robotics.111 Early exposure and positive engagement with mentors, especially role models who share diversity characteristics, is essential. However, that is often not the case. As Pereida and Greeff112 demonstrate, robotics has a diversity problem: As of 2019, "Women account for only 18% of authors at leading AI conferences, 20% of AI professorships, and 15% and 10% of research staff at Facebook and Google, respectively … black workers represent only 2.5% of Googles [sic] entire workforce and 4% of Facebooks [sic] and Microsofts [sic]. There is no data available for transgender people and other gender minorities." Involving more people of diversity is a clear next step. Initiatives like the aforementioned Black in Robotics and similar efforts, including the nonprofit Code2040113 for racial equity, the algorithmic bias-fighting Algorithmic Justice League, and the "Inclusive HRI" and "Inclusion@RSS" workshops at their respective conferences are making strides. We need to make a concerted effort to change the situation and evade the downsides of our own implicit (and perhaps other more nefarious) biases.

---

108 Caliskan, A., Bryson, J. J., and Narayanan, A., Semantics Derived Automatically from Language Corpora Contain Human-like Biases. Science 356(6334), 183–186, 2017. https://doi.org/10.1126/science.aal4230.

109 Freeing robotics and AI from bias. Science Editor's Blog. www.science.org/content/blog-post/freeing-robotics-and-ai-bias; Office of Science and Technology Policy (OSTP), Reducing the Impact of Bias in the Stem Workforce: Strengthening Excellence and Innovation. A Report of The Interagency Policy Group on Increasing Diversity in the Stem Workforce by Reducing the Impact of Bias, 2016.

What happens when robotics is not diverse? Very little work on HRI exists, but we can look to larger patterns and similar fields for answers. Joshi and Roh114 conducted a context-sensitive survey of team performance, considering social characteristics and work-related background within organizational contexts. This nuanced approach revealed that nondiverse organizational structures headed by people with more social power than those working under them lead to team performance losses. The effect was reversed when the context was different, specifically when it was not white male-led in the Western context. A more subtle form of bias can be at play, one that those benefiting from it may not realize is taking place: expertise advantage.115 Team members whose social attributes happen to match those of the

---

110 Pannier, C., Berry, C., Morris, M., and Zhao, X., Diversity and Inclusion in Mechatronics and Robotics Engineering Education. In ASEE Annual Conference Exposition Proceedings, No. 10184534, June 2020. https://par.nsf.gov/biblio/10184534; Berry (2021), supra note 109, at 26.

111 Greenemeier, L., Can Robotics Solve Its Diversity Problem? Scientific American, 2018. www.scientificamerican.com/article/can-robotics-solve-its-diversity-problem/.

112 Pereida & Greeff, Diversity in Robotics: From Diverse Teams to Diverse Impact, p. 2, 2019. www.dynsyslab.org/wp-content/papercite-data/pdf/pereida-icra19b.pdf.

113 www.code2040.org.

114 Joshi, A. and Roh, H., The Role of Context in Work Team Diversity Research: A Meta-analytic Review. Academy of Management Journal 52(3), 599–627, 2009. https://doi.org/10.5465/amj.2009.41331491.

115 Berger, J., Ridgeway, C. L., Fisek, M. H., and Norman, R. Z., The Legitimation and Delegitimation of Power and Prestige Orders. American Sociological Review, 379–405, 1998. https://doi.org/10.2307/2657555; Berger, J., leader/s will gain invisible privileges, while those whose social attributes do not match will lose out. More troubling for robotics, a survey of high-technology workplaces by DiTo-maso et al.116 found that white men in these settings were not only privileged in subtle ways, but incontrovertibly, receiving the wealth of the mentoring, training, positive performance reviews, and other resources offered by firms. These power structures likely translate to other contexts, as social power is a human attribute. Moreover, as Horwitz and Horwitz117 found in their meta-analysis, team diversity alone was not enough to change the social landscape.

We must also examine research practice. Various domains of practice have identified power imbalances when it comes to women receiving tenure and full professorships,118 being authors or first authors,119 having high citation counts at high impact venues,120 being invited as speakers and panelists,121 being deemed eligible to receive prestigious awards,122 and more. While gender is often the focus, other social characteristics and intersections may be further impacted, especially when it comes to a researcher's race, ethnicity, language,

---

Ridgeway, C. L., and Zelditch, M., Construction of Status and Referential Structures. Sociological Theory 20(2), 157–179, 2002. https://doi.org/10.1111/1467-9558.00157

116 DiTomaso, N., Post, C., and Parks-Yancy, R., Workforce Diversity and Inequality: Power, Status, and Numbers. Annual Review of Sociology 33, 473, 2007. https://doi.org/10.1146/annurev.soc.33.040406.131805

117 Horwitz, S. K. and Horwitz, I. B., The Effects of Team Diversity on Team Outcomes: A Meta-analytic Review of Team Demography. Journal of Management 33(6), 987–1015, 2007. https://doi.org/10.1177/0149206307308587

118 Jena, A. B., Khullar, D., Ho, O., Olenski, A. R., and Blumenthal, D. M., Sex Differences in Academic Rank in US Medical Schools in 2014. JAMA 314(11), 1149–1158, 2015. https://doi.org/10.1001%2Fjama.2015.10680.

119 Jagsi, R., Guancial, E. A., Worobey, C. C., Henault, L. E., Chang, Y., Starr, R., ... and Hylek, E. M., The "gender gap" in Authorship of Academic Medical Literature – a 35-year Perspective. New England Journal of Medicine 355(3), 281–287, 2006. https://doi.org/10.1056/nejmsa053910.

120 Chatterjee, P. and Werner, R. M., Gender Disparity in Citations in High-Impact Journal Articles. JAMA Network Open 4(7), e2114509–e2114509, 2021. https://doi.org/10.1001%2Fjamanetworkopen.2021.14509.

121 Fournier, L. E., Hopping, G. C., Zhu, L., Perez-Pinzon, M. A., Ovbiagele, B., McCullough, L. D., and Sharrief, A. Z., Females Are Less Likely Invited Speakers to the International Stroke Conference: Time's up to Address Sex Disparity. Stroke 51(2), 674–678, 2020. https://doi.org/10.1161%2FSTROKEAHA.119.027016.

122 Silver, J. K., Slocum, C. S., Bank, A. M., Bhatnagar, S., Blauwet, C. A., Poorman, J. A., ... and Parangi, S., Where Are the Women? The Underrepresentation of Women Physicians among Recognition Award Recipients from Medical Specialty Societies. PM&R 9(8), 804–815, 2017. https://doi.org/10.1016/j.pmrj.2017.06.001.

and/or nationality.123 Lerback124 found that the US may be particularly WEIRD,125 with paper acceptance rates and citations being negatively impacted by diverse authorship … but only in the States. Globally, diverse teams increased the impact of the research published. Little is known about the state of affairs in HRI. Baxter et al.126 found that about half of recent HRI work sampled from university populations, which are largely WEIRD or EIRD; that is, made of participants from Western (or perhaps non-Western), educated, industrialized, rich, and democratic nations. Seaborn and Frank127 found implicit biases in the way gender was framed for Pepper, a famous and widely used humanoid robot. Findings pointed to subtle influences on participant responses based on the researchers' stated and unstated formulations of gender. More work is needed to identify to what extent such biases play out in HRI and robotics work, on "our" side, but we are only human – we should not be surprised to find that we are just as biased as everyone else.

The good news: As a still-new field, social robotics and the subfield of HRI are well positioned to avoid the problems still entrenched in older fields of study.128 We would do well to think of robotics as a social practice, not only because of the social robots and human subjects research, but because science is social.129 We need to shake up the team as well as who is leading it. We need to recognize how social power operates within robotics practice. Indeed, as Leslie et

---

123 Kwon, D., The Rise of Citational Justice: How Scholars are Making References Fairer. Nature 603(7902), 568–571, 2022. https://doi.org/10.1038/d41586-022-00793-1.

124 Lerback, J. C., Hanson, B., and Wooden, P., Association between Author Diversity and Acceptance Rates and Citations in Peer-Reviewed Earth Science Manuscripts. Earth and Space Science 7(5), e2019EA000946, 2020. https://doi.org/10.1029/2019EA000946.

125 Henrich, J., Heine, S. J., and Norenzayan, A., The Weirdest People in the World? Behavioral and Brain Sciences 33(2–3), 61–83, 2010. https://doi.org/10.1017/S0140525X0999152X.

126 Baxter, P., Kennedy, J., Senft, E., Lemaignan, S., and Belpaeme, T., From Characterising Three Years of HRI to Methodology and Reporting Recommendations. In 2016 11th ACM/IEEE International Conference on Human-Robot Interaction (HRI), pp. 391–398, March 2016. IEEE. https://doi.org/10.1109/HRI.2016.7451777.

127 Seaborn and Frank (2022), supra note 6, at 3.

128 Howard, A. and Borenstein, J., The Ugly Truth about Ourselves and our Robot Creations: The Problem of Bias and Social Inequity. Science and Engineering Ethics 24(5), 1521–1536, 2018. https://doi.org/10.1007/s11948-017-9975-2.

129 Nature Genetics, Science Is Social. Nature Genetics 50(1619), 2018. https://doi.org/10.1038/s41588-018-0308-4.

al.130 recently discovered, taking an "identity-blind" approach can lead to spurious outcomes. However, embracing a multicultural perspective on team diversity that recognizes the importance of social power typically leads to an array of positive intergroup outcomes for all involved. When teams become diverse, everyone benefits.

I offer the following antibiasing provocations and modes of reflexivity for the reader to peruse and pursue:

**Critical next steps:**

- We can critically reflect on our team's diversity and lack thereof.
- We can seek out colleagues and collaborators who are different from ourselves. We do not have to stay within our organizational or national boundaries, especially in the post-COVID-19 future.
- We can give those with less social power in the team access to other forms of power, such as leadership positions, and resources – these are often not considered yet crucial to each person's success.
- We can conduct surveys and other forms of research to assess the degree to which diversity has been achieved in robotics research and practice, and how this diversity has influenced our work.
- We can get involved with diversity organizations and training initiatives.
- We can recognize how power operates within community structures, including academic ones. For instance, we can cite a diversity of authors rather than the most famous ones. We can use our power to give overlooked research a boost.

**Social factors:**

- We must be aware of power structures and intersectionality within the team, not only in terms of role differences in nonflat structures, but how social power can operate to create differences in power and unwelcoming, or worse, environments.

**Legal factors:**

- We can cocreate and circulate regulations about harassment. We must be clear about the repercussions for harassment, exclusion, and so on. If a team member harasses another team member, will they agree on how "harassment" is defined, and how it should be dealt with? These structures need to be in place and agreed upon by all parties before any possible harassment can occur.

---

130 Leslie, L. M., Bono, J. E., Kim, Y. S., and Beaver, G. R., On Melting Pots and Salad Bowls: A Meta-Analysis of the Effects of Identity-Blind and Identity-Conscious Diversity Ideologies. Journal of Applied Psychology 105(5), 453, 2020. https://psycnet.apa.org/doi/10.1037/apl0000446.

- We must enforce these regulations. If you find that you cannot, seek outside help from **objective third parties with experience in diversity.**

**Ethical factors:**

- We must avoid performative gestures of diversity. We should not just hire underpowered folks but also share what power we have with them.
- We can encourage our collaborators and institutions to share diversity data openly. We can then target areas that need representation.
- We can listen to those marked by diversity and follow through on critical feedback. We should not rely on them to make necessary changes.
- We should engage social peers for self-checks rather than rely on the labor of those with less social power in the team.

## 19.4  Conclusion

Bias cannot be eradicated, but it does not need to be. We only need to recognize it, understand it, and be conscious of how we do or do not address it. Taking a critical lens to the socially expressive humanoid robots that we work on, study, and increasingly live with is fruitful for understanding how biases crop up, as well as how they might be addressed. I have outlined a series of cases along these parallel lines that I hope demonstrates where we are at – and where we can go from here. The social, ethical, and legal repercussions should incite curiosity and a little trepidation. We can move forward with purpose and community-level reflexivity, continuing initial efforts through training camps, workshops, edited volumes and calls, and academic–industrial partnerships. Future critical review work, especially systematic reviews and meta-analyses targeting certain forms of bias will show just how far we have progressed. New laws, notions tested in court, and ethically aware forms of social robots and interactions will demonstrate whether and how we have made headway. The critical next steps outlined above can be explored across a wide array of humanlike and socially aware robots and HRI contexts. We have made interactive humanoid robots in our image; now we can explore the repercussions and opportunities of this with principled intent.

## Acknowledgments

I wish to thank Kristiina Jokinen for being a part of formative discussions on this chapter, as well as providing feedback on and editorial tweaks to an earlier draft. I would also like to thank Jacqueline Urakami, Hiroki Oura, and all participants of BoAB 2021, who inspired and motivated this work. I also thank Peter Pennefather for ongoing critical engagements on the topics of bias, robots, and humanity that inspired several of the ideas I have explored here.